\documentclass[runningheads]{llncs}
\usepackage{graphicx}
\usepackage{tikz}
\usepackage{comment}
\usepackage{amsmath,amssymb} % define this before the line numbering.
\usepackage{color}
\usepackage{orcidlink}

\usepackage[accsupp]{axessibility}  % Improves PDF readability for those with disabilities.

\hypersetup{hidelinks}
\usepackage[misc]{ifsym}
\usepackage{multirow}
\usepackage{diagbox}
\usepackage{adjustbox}
\usepackage[caption=false, font=scriptsize]{subfig}
\newcommand{\ie}[1]{{\textit{i.e.}{{#1}}}}
\newcommand{\eg}[1]{{\textit{e.g.}{{#1}}}}
\newcommand{\mt}[1]{{\noindent\textbf{{#1}}}}
\newcommand{\mkr}[1]{{\color{red}{{#1}}}}
\newcommand{\mkg}[1]{{\color{green}{{#1}}}}
\newcommand{\mkb}[1]{{\color{blue}{{#1}}}}

\begin{document}
% \linenumbers
\pagestyle{headings}
\mainmatter
\def\ECCVSubNumber{1676}  % Insert your submission number here

\title{Hierarchical Feature Alignment Network for Unsupervised Video Object Segmentation} % Replace with your title

\begin{comment}
\titlerunning{ECCV-22 submission ID \ECCVSubNumber} 
\authorrunning{ECCV-22 submission ID \ECCVSubNumber} 
\author{Anonymous ECCV submission}
\institute{Paper ID \ECCVSubNumber}
\end{comment}
%******************

%\begin{comment}
\titlerunning{Hierarchical Feature Alignment Network for UVOS}
\author{Gensheng Pei\inst{1}\orcidlink{0000-0002-7677-7487} \and
Fumin Shen\inst{2}\textsuperscript{(\Letter)}\orcidlink{0000-0001-7303-3231} \and 
Yazhou Yao\inst{1}\orcidlink{0000-0002-0337-9410} \and
Guo-Sen Xie\inst{1}\textsuperscript{(\Letter)}\orcidlink{0000-0002-5487-9845} \and \\
Zhenmin Tang\inst{1}\orcidlink{0000-0001-6708-2205} \and
Jinhui Tang\inst{1}\orcidlink{0000-0001-9008-222X}
}
\authorrunning{G. Pei et al.}
\institute{Nanjing University of Science and Technology, Nanjing, China \\
\email{yazhou.yao@njust.edu.cn} \and
University of Electronic Science and Technology of China, Chengdu, China\\
\url{https://github.com/NUST-Machine-Intelligence-Laboratory/HFAN}
}
%\end{comment}
%******************

\maketitle

\begin{abstract}
Optical flow is an easily conceived and precious cue for advancing unsupervised video object segmentation (UVOS). Most of the previous methods directly extract and fuse the motion and appearance features for segmenting target objects in the UVOS setting. However, optical flow is intrinsically an instantaneous velocity of all pixels among consecutive frames, thus making the motion features not aligned well with the primary objects among the corresponding frames. To solve the above challenge, we propose a concise, practical, and efficient architecture for appearance and motion feature alignment, dubbed hierarchical feature alignment network (HFAN). Specifically, the key merits in HFAN are the sequential \textbf{F}eature \textbf{A}lign\textbf{M}ent (FAM) module and the \textbf{F}eature \textbf{A}dapta\textbf{T}ion (FAT) module, which are leveraged for processing the appearance and motion features hierarchically. FAM is capable of aligning both appearance and motion features with the primary object semantic representations, respectively. Further, FAT is explicitly designed for the adaptive fusion of appearance and motion features to achieve a desirable trade-off between cross-modal features. Extensive experiments demonstrate the effectiveness of the proposed HFAN, which reaches a new state-of-the-art performance on DAVIS-16, achieving 88.7 $\mathcal{J}\&\mathcal{F}$ Mean, \ie, a relative improvement of 3.5\% over the best published result.
\keywords{Video object segmentation \and Feature alignment}
\end{abstract}

\section{Introduction}
Video object segmentation (VOS) aims to segment objects for each frame in a video sequence. 
Compared to semi-supervised VOS (SVOS), in which annotations are provided for the first frame at test time, unsupervised VOS (UVOS) is particularly challenging as it involves no prior knowledge and human interposing. This work focuses on the UVOS task, which has motivated numerous downstream segmentation topics~\cite{MiVOS,TAODA,FGVI,I2CRC,NSROM}.

UVOS approaches can be grouped into three main subcategories: motion-based, appearance-based, and motion-appearance-based, depending on the utilizations of different feature types. By merely using motion information~\cite{FOSUV,LSMO}, the UVOS is transformed into a moving object segmentation (MOS) task. The main drawback of MOS is the risk of losing targets when the object is moving slowly or is stationary. Further, appearance-based methods~\cite{AGNN,3DCSeg,F2Net,DFNet} usually describe the target in detail using mature image segmentation techniques. However, the lack of prior knowledge on the primary objects in the unsupervised solution, almost always, can lead to mis-segmentation cases. By contrast, motion-appearance schemes~\cite{EPO+,AMC-Net,TransportNet,RTNet,MATNet} can mitigate the deficiencies of the above two types of methods. Appearance features compensate for the shortage of motion descriptions on semantic representations, and motion cues enable the high-quality candidate regions to be selected for appearance features. 

As the leading motion-appearance scenario, optical flow guided UVOS methods have significantly advanced the performances of segmentation. Abandoning appearance modeling by converting VOS into a foreground motion prediction based purely on optical flow information does not handle static foreground objects well. However, two intrinsic drawbacks exist in these approaches. \textbf{First}, optical flow describes the velocity vector of per-pixel motion in a video, capturing the motion information between consecutive frames. As such, the positions of the primary objects in a frame and/or its corresponding optical flow are usually not well preserved. Most of the existing methods solely fuse the features of a frame and its corresponding optical flow without considering the respective alignments between the primary objects and the appearance/motion features. This inevitably leads to the loss of boundary information of the primary objects. \textbf{Second}, when facing occlusions, motion blur, fast-moving objects, and even stationary objects in UVOS, the inferior optical flow estimation directly affects the segmentation results of the final models. Especially, if optical flow estimation fails, the motion features of the primary objects in the video are invalidated accordingly. In this case, the unselective fusion of appearance and motion features is probably harmful to segmentation accuracy. 

\begin{figure}[t]
	\centering
	\subfloat{\includegraphics[width=0.49\linewidth]{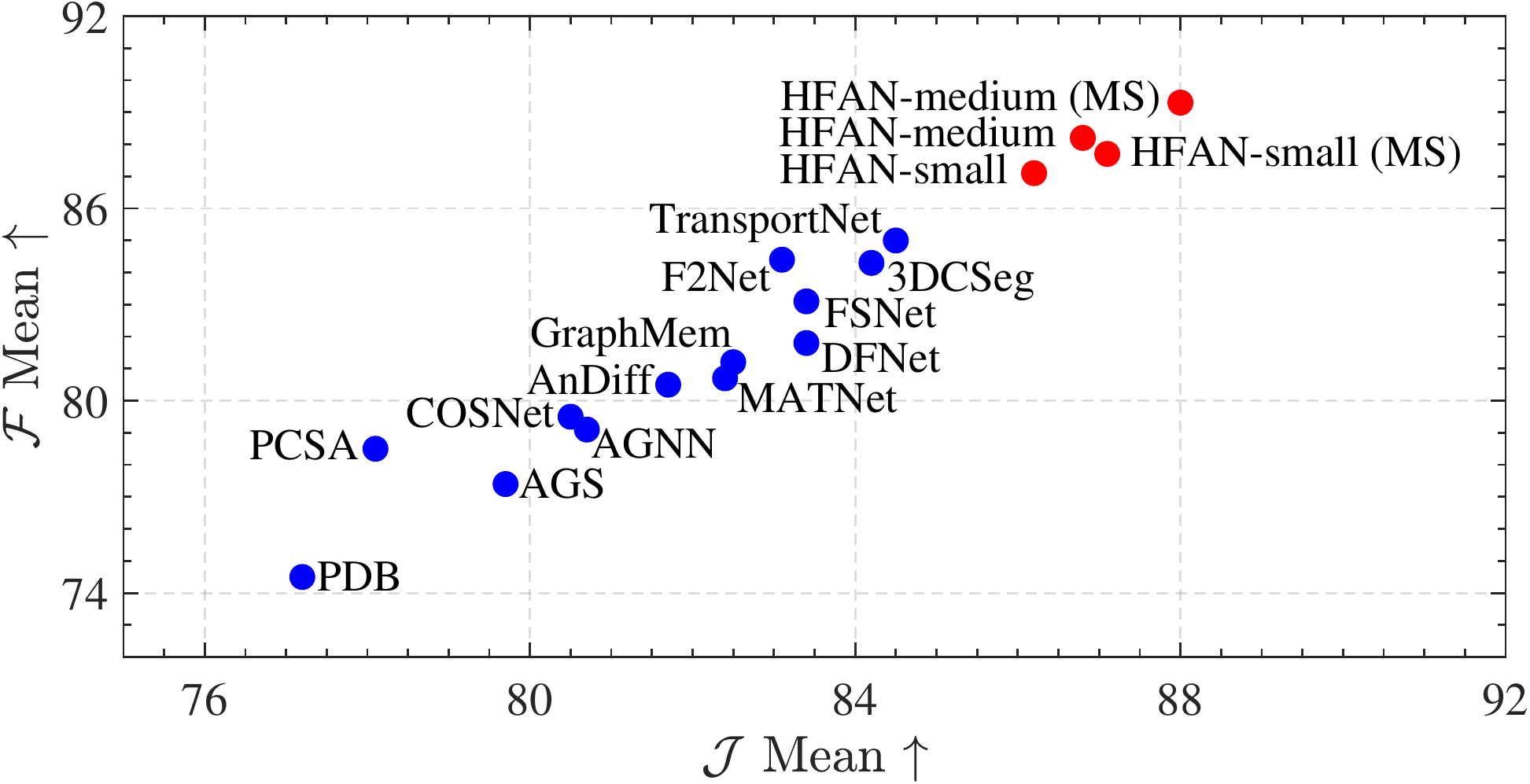}}
	\hfill
	\subfloat{\includegraphics[width=0.49\linewidth]{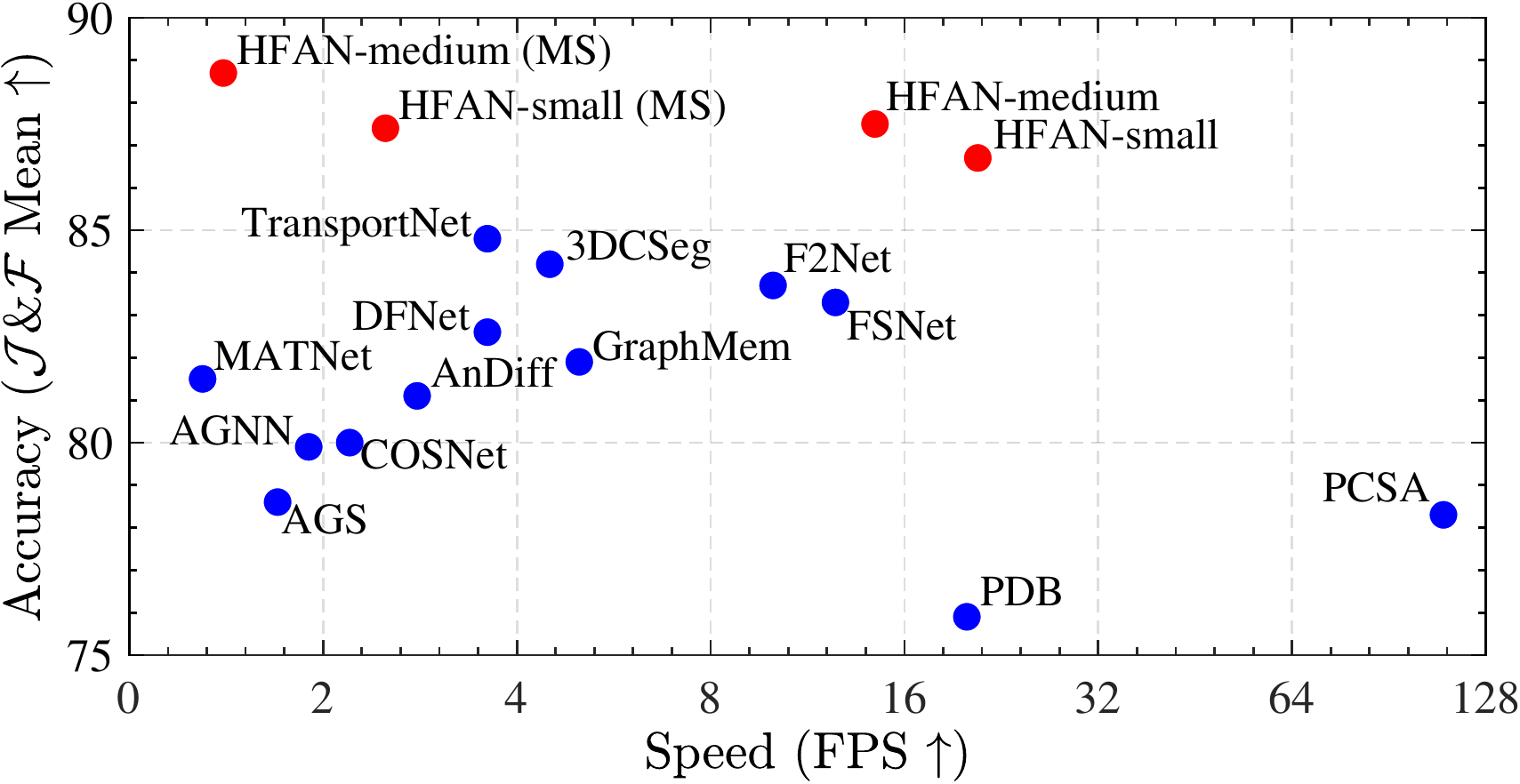}}
	\caption{Performance $\mathcal{J}\&\mathcal{F}$ Mean versus inference speed FPS (frames per second) on DAVIS-16~\cite{DAVIS-16}. Existing and proposed methods are marked with {\large\textcolor{blue}{$\bullet$}} and {\large\textcolor{red}{$\bullet$}}, respectively. }
	\label{fig:speed_acc}
\end{figure}

To tackle the above challenges, inspired by the current trend of optical flow guided UVOS~\cite{LSMO,EPO+,FSNet,AMC-Net,TransportNet,RTNet,MATNet}, we propose a hierarchical feature alignment network (HFAN). It aligns the object positions with the motion/appearance features and adapts the aligned features for mitigating the cross-modal mismatch. Specifically, we construct a \textbf{F}eature \textbf{A}lign\textbf{M}ent (FAM) module to implement object-level alignment with appearance/motion features in the multi-level feature encoding stage. Considering that the spatial locations of appearance and ground-truth target regions are seamlessly matched, we generate the coarse segmentation probability mask by appearance features. Next, FAM leverages the same object regions (\ie, coarse masks) to represent object-level alignment features for appearance and motion. Moreover, we build the \textbf{F}eature \textbf{A}dapta\textbf{T}ion (FAT) module to combine appearance and motion features after the alignment step. FAT aims to ensure the robustness of fused features by constructing an adaptive weight between appearance and motion features. Notably, the adaptive fusion of appearance and motion features could effectively relieve the harms of optical flow estimation failure and motion blur on segmentation results.

We assess the effectiveness and reliability of the proposed model on three widely-used datasets. On DAVIS-16~\cite{DAVIS-16}, our HFAN-small and HFAN-medium achieve 86.7 and 87.5 $\mathcal{J}\&\mathcal{F}$ Mean, respectively, at 20.8 and 14.4 FPS. These are new state-of-the-art (SOTA) results in terms of accuracy and speed, as shown in Fig.~\ref{fig:speed_acc}. On YouTube-Objects~\cite{YouTube-Objects}, the proposed HFAN-small represents a relative improvement of 2.0\% over the reported best result. Furthermore, the proposed method achieves an equivalent performance to SVOS models on Long-Videos~\cite{AFB-URR}. Meanwhile, HFAN also reaches the best reported result on DAVIS-16 for the video salient object detection (VSOD), which aims to detect salient regions in videos. In summary, HFAN provides an efficient solution and a new perspective on optical flow guided UVOS.

\section{Related Work}
\subsection{Video Object Segmentation}
Current video object segmentation is broadly classified as unsupervised VOS \cite{AGNN,AGS,AnDiff,LVO} and semi-supervised VOS~\cite{JOINT,STM,HMMN,SSTVOS} tasks. The main difference is whether they provide accurate pixel-level annotations for the first frame of the segmented video at inference. As research on VOS has progressed, interactive VOS methods~\cite{IVOS,GIVOS,MiVOS} that utilize user interaction (\eg, scribbles or clicks) as input to iteratively optimize segmentation results have yielded good performance. Referring VOS setting~\cite{CSTM,URVOS,RefVOS} arises from considering a different type of interaction, language-guided video expressions, \ie, target objects referred by a given language descriptions. However, the expensive nature of high-quality annotated video data motivates the need for an elegant and unrestricted VOS setting. This paper focuses on UVOS, which does not use any human intervention during testing. Depending on whether the current methods use deep features or not, we further divide UVOS into two subcategories: \textit{traditional} and \textit{deep}.

The computer vision community has extensively studied the task of automatic binary video segmentation over nearly three decades. Early \textit{traditional} models were typically based on specific heuristics related to the foreground (\ie, target proposals~\cite{FCOP}, motion boundaries~\cite{FOSUV}, salient objects~\cite{SAG}). They required hand-crafted low-level features (\eg, SIFT, edges). Later, several methods (\eg, point trajectories~\cite{SMO}, background subtraction~\cite{DMFBS} and over-segmentation~\cite{POLP}) were proposed to segment and track all targets with different motions and appearances in the video. More recently, with the renaissance of artificial neural networks, \textit{deep} models (\eg, CNN~\cite{DFNet,AnDiff}, RNN~\cite{PDB,AGS,EPO+}, GNN~\cite{AGNN,GraphMem}) have enabled UVOS to evolve rapidly. A quintessential example of an attempt to apply deep learning techniques in this field is LSMO~\cite{LSMO}, which learns a multilayer perceptron to detect moving objects. The computational burden is reduced by many subsequent approaches based on fully convolutional networks, such as two-stream structures~\cite{LVO,Fusionseg,MGA,MotAdapt}, CNN-based encoder-decoder architectures~\cite{MATNet,Segflow,TAODA}, and Siamese network~\cite{COSNet,F2Net}. As the field of optical flow estimation~\cite{FlowNet-2.0,RAFT,PWCNet} has flourished, more and more optical flow based UVOS methods~\cite{FSNet,AMC-Net,TransportNet,RTNet} have gained tremendous performance improvements. The major difference from the optical flow-based approaches described above is that we reconsider the mismatch between frames and optical flow. Our HFAN performs hierarchical feature alignment and adaptation of motion-appearance features to achieve accurate feature representation of the primary objects in a video.

\subsection{Feature Alignment}
Feature alignment is widely used in various fields, \eg, object detection~\cite{AlignDet,S2ANet,FPN}, image segmentation~\cite{FaPN,SFNet,OCR,AlignSeg}, and person re-identification~\cite{EDI,PGFA,JPFA,Jo-SRC}. For object detection, feature alignment mainly involves the misalignment between anchor boxes and convolutional features, in addition to multiple anchors for the same point in the feature map. Existing image segmentation models generally adopt the feature pyramid networks (FPN)~\cite{FPN} to obtain different resolution feature maps to improve performance. However, this increases the loss of boundary information during downsampling and unaligned feature maps with different resolutions for upsampling. An effective way~\cite{FaPN,SFNet} is to align the features from the coarsest resolution to the finest resolution to match positions between feature maps. Aligning and adapting motion and appearance features of multi-level representations from the same encoder is implemented by our HFAN. Thus, it is guaranteed that hierarchical feature maps between the two modalities align their respective features based on the same primary objects. 

\section{The Proposed Method}
Our HFAN consists of two modules: \textit{feature alignment} (FAM, \S\ref{FAM}) and \textit{feature adaptation} (FAT, \S\ref{FAT}). FAM aligns the hierarchical features of appearance and motion feature maps with the primary objects. FAT fuses these two aligned feature maps at the pixel-level with a learnable adaptive weight.

\begin{figure*}[t]
	\centering
	\includegraphics[width=\linewidth]{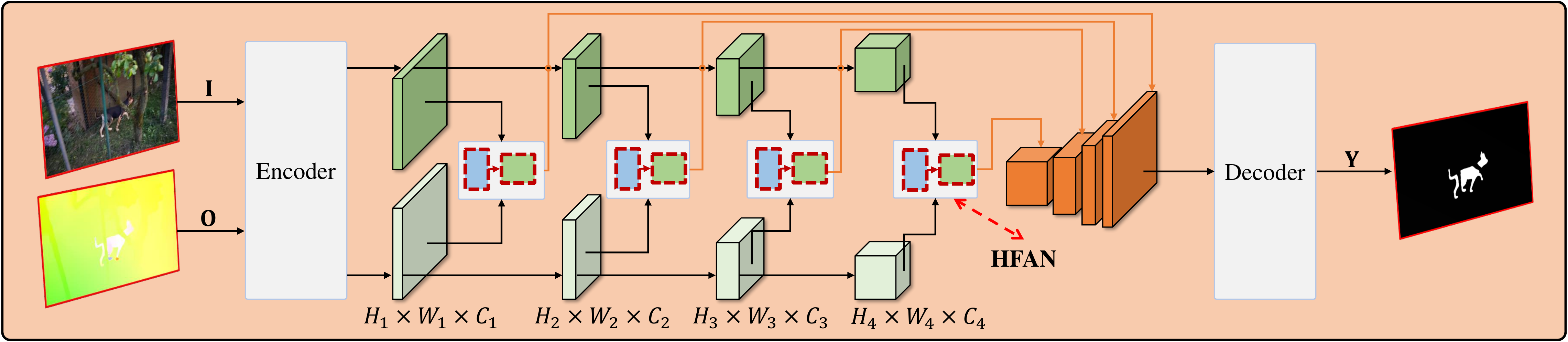}
	\caption{\textbf{The pipeline of HFAN}. Frame $\mathbf{I}$ and optical flow $\mathbf{O}$ are used as inputs to extract hierarchical appearance and motion features, respectively, through an encoder with HFAN. And the excellent segmentation mask $\mathbf{Y}$ is obtained by the decoder.}
	\label{fig:pipeline}
\end{figure*}

\subsection{Task Definitions}
Given an input video $\mathcal{I}$ with $N$ frames, we can select each frame $\mathbf{I}\in\mathbb{R}^{H\times{W}\times{3}}$, and calculate the relative optical flow $\mathbf{O}\in\mathbb{R}^{H\times{W}\times{3}}$ (visualized as an RGB image) by~\cite{RAFT}. In the $i$-th stage of the multi-level feature representation ($i\in\{1,2,3,4\}$), the appearance and motion features are denoted as $\mathbf{I}_{i}\in\mathbb{R}^{{H_i}\times{W_i}\times{C_i}}$ and $\mathbf{O}_{i}\in\mathbb{R}^{{H_i}\times{W_i}\times{C_i}}$, respectively. ${H_i}\times{W_i}$ indicates the feature resolution, where the value is set to ${\frac{H}{2^{i+1}}}\times{\frac{W}{2^{i+1}}}$. The proposed HFAN aims to generate object-level aligned, high-quality adapted features,
\begin{equation}\label{Eq.1}
	\begin{split}
		\mathbf{U}_i=\mathcal{F}_{\rm{HFAN}}(\mathbf{I}_{i},\mathbf{O}_{i})\in\mathbb{R}^{{H_i}\times{W_i}\times{C_i}}.
	\end{split}
\end{equation}
Here, $\mathcal{F}_{\rm{HFAN}}(\cdot,\cdot)$ contains two main modules, which are:
\begin{equation}\label{Eq.2}
	\begin{split}
		\mathbf{\hat{I}}_{i},\mathbf{\hat{O}}_{i}=\mathcal{F}_{\rm{FAM}}(\mathbf{I}_{i},\mathbf{O}_{i}),
		\mathbf{U}_i=\mathcal{F}_{\rm{FAT}}(\mathbf{\hat{I}}_{i},\mathbf{\hat{O}}_{i}),
	\end{split}
\end{equation}
where $\mathcal{F}_{\rm{FAM}}$ conducts feature alignment for $\mathbf{I}_i$ and $\mathbf{O}_i$, and $\mathcal{F}_{\rm{FAT}}$ fuses aligned feature maps from $\mathcal{F}_{\rm{FAM}}$ by performing a multi-modal adaptive feature fusion. The overall architecture of the proposed method is shown in Fig.~\ref{fig:pipeline}. 

We adopt a lightweight MiT~\cite{SegFormer} backbone (ResNet~\cite{ResNet} is also studied, see \S\ref{subsec:Ablation} for details.) and employ a decoder to yield the primary object binary mask $\mathbf{Y}\in\{0,1\}^{{H}\times{W}}$ of the frame $\mathbf{I}$. Next, we illustrate in detail the two main modules of our proposed HFAN model, along with the training and inference phases.

\subsection{Feature Alignment Module}\label{FAM}
Optical flow methods produce a dense motion vector field by generating a vector for each pixel, which is important auxiliary information for studying video analysis and representation. Former works~\cite{MATNet,AMC-Net,TransportNet,RTNet} using optical flow guidance have defaulted one video frame and its optical flow to aligned images. However, this hypothesis then holds approximately only if the motion between two consecutive frames is small. In addition, this solidification tends to result in poor accuracy along moving object boundaries. An intuitive concept is that although appearance features and motion features are unaligned, the bond between them is that they share the primary objects. Motivated by this, we design a feature alignment module specifically for the frame and optical flow to alleviate these issues. Firstly, FAM predicts the coarse segmentation probability of $\mathbf{I}_i$ to obtain
\begin{equation}\label{Eq.3}
	\begin{split}
		\mathbf{P}_{i}&=\mathcal{F}_{\rm{CS}}(\mathbf{I}_{i})\in\mathbb{R}^{{H_i}\times{W_i}\times{N_{cls}}},
	\end{split}
\end{equation}
where $\mathcal{F}_{\rm{CS}}(\cdot)$ represents the coarse segmentation probability mask implemented by a convolution block \texttt{${\texttt{Conv}}_{{1}\times{1}}(C_{i},N_{cls})\rightarrow\texttt{BN}\rightarrow\texttt{ReLU}$} on the appearance feature map $\mathbf{I}_i$, and $N_{cls}$ denotes the number of categories. Here, \texttt{BN} indicates the batch normalization~\cite{BatchNorm} and \texttt{ReLU} is the rectified linear unit~\cite{ReLU}. This paper focuses on single foreground and background, so $N_{cls}$ is set to 2. 

The regions contained in the coarse probability mask $\mathbf{P}_i$ obtained by the original frame $\mathbf{I}_i$ are consistent with the primary object's areas to be segmented. Therefore, we design the feature alignment module, which only aligns the appearance and motion features separately for the mask regions. This way has the merit of reducing the computational cost while weakening the negative impact of the optical flow background noise on the segmentation. Subsequently, $\mathbf{P}_i$ obtained by Eq.~\eqref{Eq.3} is a contextual representation of primary object regions co-built with the original appearance feature map. We design the category-specific semantic (CSS) module to represent the category semantic, formulated as
\begin{equation}\label{Eq.4}
	\begin{split}
		\mathbf{I}_{i}^{'}=\texttt{permute}(\texttt{view}(\mathbf{I}_{i})), \mathbf{P}_{i}^{'}=\texttt{softmax}(\texttt{view}(\mathbf{P}_{i})), \\
		\mathbf{M}_{i}=\mathcal{F}_{\rm{CSS}}(\mathbf{I}_{i},\mathbf{P}_{i})=\texttt{matmul}(\mathbf{I}_{i}^{'},\mathbf{P}_{i}^{'})\in\mathbb{R}^{{C_i}\times{N_{cls}}\times{1}}, \\
	\end{split}
\end{equation}
where \texttt{view}, \texttt{permute} and \texttt{matmul} indicate the reshaping, permuting tensor dimension and tensor product. \texttt{softmax} is also known as normalized exponential function. The prominent role of $\mathbf{M}_i$ is summarized in two points: \textbf{1)} the spatial compression of appearance features within a specific region $\mathbf{P}_i$; \textbf{2)} the construction of category-specific information shared by appearance and motion features. The interchange on feature and semantic levels performed by $\mathcal{F}_{\rm{CSS}}$ makes it possible to seek common contexts for appearance-motion features.

Immediately afterward, the primary object context (POC) module is devised to perform object-level contextual alignment of appearance and motion features with the same $\mathbf{M}_{i}$. Inspired by self-attention~\cite{Attention}, $\mathcal{F}_{\rm{SA}}$ is achieved by
\begin{equation}\label{Eq.5}
	\begin{split}
		\mathcal{F}_{\rm{SA}}&=\texttt{softmax}(\alpha\mathbf{Q}\mathbf{K}^{\rm{T}})\mathbf{V}, \mathbf{Q}\in\{\mathbf{I}_{i},\mathbf{O}_{i}\}\\
		\mathbf{Q},\mathbf{K},\mathbf{V}&=\mathcal{F}_{\rm{Query}}(\mathbf{Q}),\mathcal{F}_{\rm{Key}}(\mathbf{M}_{i}),\mathcal{F}_{\rm{Value}}(\mathbf{M}_{i}),
	\end{split}
\end{equation}
where $\mathbf{Q}$, $\mathbf{K}$ and $\mathbf{V}$ denote the query, key and value obtained by three transformation operations of $\mathcal{F}_{\rm{Query}}$, $\mathcal{F}_{\rm{Key}}$ and $\mathcal{F}_{\rm{Value}}$, respectively. They are formed by ${\texttt{Conv}}_{{1}\times{1}}(C_{i},C_{i}/r)\rightarrow\texttt{BN}\rightarrow\texttt{ReLU}$. $\alpha=\frac{1}{\sqrt{C_i}}$ is a scaling factor. $r$ is set to $C_i/16$ for channel reduction ratio and \texttt{concat} indicates the concatenation operation.

Our proposed POC module helps refine the target boundaries and alleviate the primary object shifts between the frame and optical flow. The POC module can be represented as follows:
\begin{equation}\label{Eq.6}
	\begin{split}
		\mathbf{\hat{I}}_{i}&=\mathcal{F}_{\rm{POC}}(\mathbf{I}_{i},\mathbf{M}_{i})\in\mathbb{R}^{{H_i}\times{W_i}\times{C_i}}, \\
		\mathbf{\hat{O}}_{i}&=\mathcal{F}_{\rm{POC}}(\mathbf{O}_{i},\mathbf{M}_{i})\in\mathbb{R}^{{H_i}\times{W_i}\times{C_i}},
	\end{split}
\end{equation}
where $\mathbf{\hat{I}}_{i}$ and $\mathbf{\hat{O}}_{i}$ are appearance-aligned and motion-aligned feature maps, respectively.

\begin{figure*}[t]
	\centering
	\includegraphics[width=\linewidth]{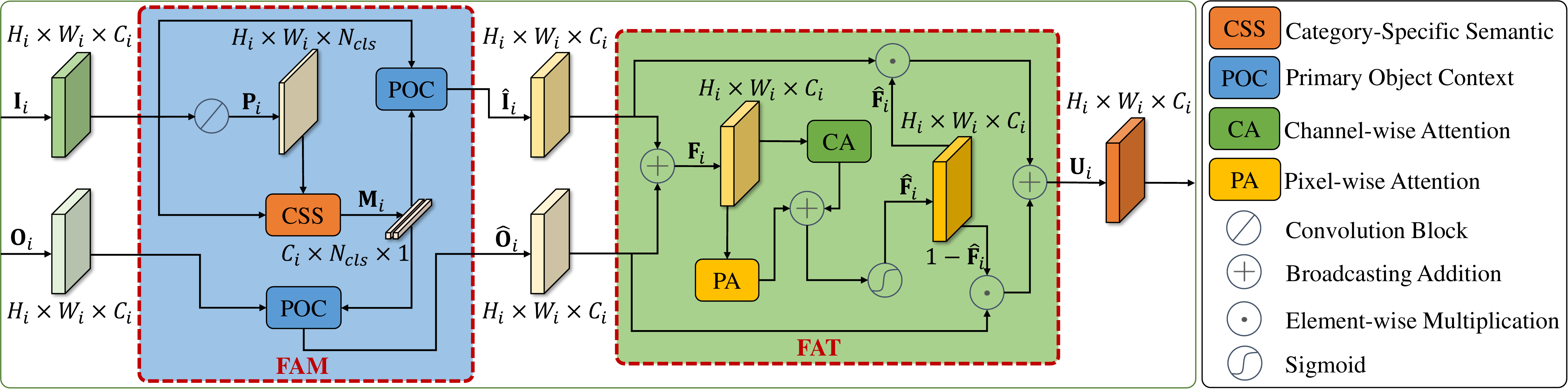}
	\caption{Illustration of the proposed FAM and FAT modules. \textit{Feature Alignment} and \textit{Feature Adaptation} are applied on each hierarchical feature map to resolve the position and modal mismatches between optical flow and video frames. $\mathbf{U}_i$ denotes the alignment and adaptation features of stage $i\in\{1,2,3,4\}$.}
	\label{fig:module}
\end{figure*}

Compared to previous methods, FAM does not interact directly with appearance and motion features but employs CSS and POC modules to achieve contextual alignment of different modal features.
As shown in Fig.~\ref{fig:module}-FAM, when $\mathbf{I}_i$ and $\mathbf{O}_i$ go through FAM, their respective features represent the shared primary object region $\mathbf{P}_i$, guided by $\mathbf{M}_i$. It ensures the feature independence of appearance $\mathbf{\hat{I}}_{i}$ and motion $\mathbf{\hat{O}}_{i}$ before the feature adaptation fusion phase.

\subsection{Feature Adaptation Module}\label{FAT}
After the appearance and motion features are expressed based on the same object-contextual region, the aligned feature maps $\mathbf{\hat{I}}_{i}$ and $\mathbf{\hat{O}}_{i}$ have more boundary information and less background noise. However, when the optical flow estimation fails due to slow motion or stationary target objects, retaining all the optical flow features would cause a tremendous loss in segmentation performance. To this end, we require adaptive operations between cross-modal features for them. In this work, we propose the feature adaptation (FAT) module. 

Specifically, we aggregate appearance-aligned and motion-aligned features and accordingly get the fused feature map $\mathbf{F}_i$, which embraces all the information of $\mathbf{\hat{I}}_{i}$ and $\mathbf{\hat{O}}_{i}$. The formula is directly expressed as $\mathbf{F}_{i}=\mathbf{\hat{I}}_{i}+\mathbf{\hat{O}}_{i}.$
Here, $\mathbf{F}_{i}\in\mathbb{R}^{{H_i}\times{W_i}\times{C_i}}$ is treated as a semantic feature map after the superposition of appearance and motion contexts, equivalent to performing a skip connection operation~\cite{ResNet,FPN} on different modal features of the same resolutions. Inspired by~\cite{SKNet}, channel-level and pixel-level semantic representations are obtained by
\begin{equation}\label{Eq.8}
	\begin{split}
		\mathbf{F}_{i}^{\rm{CA}}&=\mathcal{F}_{\rm{CA}}(\mathbf{F}_{i})\in\mathbb{R}^{{H_i}\times{W_i}\times{C_i}}, \\
		\mathbf{F}_{i}^{\rm{PA}}&=\mathcal{F}_{\rm{PA}}(\mathbf{F}_{i})\in\mathbb{R}^{{1}\times{1}\times{C_i}},
	\end{split}
\end{equation}
where $\mathcal{F}_{\rm{CA}}(\cdot)$ and $\mathcal{F}_{\rm{PA}}(\cdot)$ indicate channel-wise and pixel-wise attention operations are performed on $\mathbf{F}_{i}$.

Instead of using $\mathbf{F}_{i}$ directly as the fused appearance and motion features as in existing approaches~\cite{RTNet,MATNet,Fusionseg}, we propose to adapt these features. Specifically, we transform $\mathbf{F}_{i}$ into basis weight with feature adaptation to ensure stable feature representation capability even under low-quality motion information conditions (\eg, occlusion and slow motion). The formula is expressed as
\begin{equation}\label{Eq.9}
	\begin{split}
		\mathbf{\hat{F}}_{i}&=\texttt{sigmoid}(\mathbf{F}_{i}^{\rm{CA}}+\mathbf{F}_{i}^{\rm{PA}})\in\mathbb{R}^{{H_i}\times{W_i}\times{C_i}}, \\
		\mathbf{U}_i&=\mathbf{\hat{I}}_{i}\odot\mathbf{\hat{F}}_{i}+ \mathbf{\hat{O}}_{i}\odot(1-\mathbf{\hat{F}}_{i})\in\mathbb{R}^{{H_i}\times{W_i}\times{C_i}},
	\end{split}
\end{equation}
where $\odot$ denotes the element-wise multiplication. At this point, details of FAT are introduced, and the workflow is illustrated in Fig.~\ref{fig:module}-FAT. Further observation of Eq.~\eqref{Eq.9} shows that when $(1-\mathbf{\hat{F}}_{i})$ approaches 0, all information of $\mathbf{U}_i$ is provided by appearance features, while when $\mathbf{\hat{F}}_{i}$ reaches 0, all information of $\mathbf{U}_i$ is supplied by motion features. Meanwhile, $\mathbf{\hat{F}}_{i}$ is learnable, so it realizes the feature self-adaptation of the frame and optical flow.

\subsection{Training and Inference}
The multi-level features $\mathbf{U}_{i}$ ($i\in\{1,2,3,4\}$) obtained through HFAN are fed to the decoder $\mathcal{F}_{\rm{DEC}}$, and the predicted segmentation mask $\mathbf{Q}$ is acquired
\begin{equation}\label{Eq.10}
	\begin{split}
		\mathbf{Q}&=\mathcal{F}_{\rm{DEC}}(\mathbf{U}_{i|i=1,2,3,4})\in\mathbb{R}^{{H}\times{W}\times{N_{cls}}},
	\end{split}
\end{equation}
where $\mathcal{F}_{\rm{DEC}}(\cdot)$ utilizes a lightweight All-MLP decoder provided by~\cite{SegFormer} to ensure consistency with the encoder network MiT.

Our model is trained to minimize the loss function $\mathcal{L}$ as follows
\begin{equation}\label{Eq.11}
	\begin{split}
		\mathcal{L}=\frac{1}{H\times{W}}\sum_{p,q}\mathcal{L}_{\rm{CE}}(\mathbf{Q}_{[\cdot,p,q]},\mathbf{G}_{[p,q]}),
	\end{split}
\end{equation}
where $\mathcal{L}_{\rm{CE}}$ is the Cross Entropy Loss. $\mathbf{G}$ stands for the ground-truth mask. $\sum_{p,q}$ denotes the sum over all positions on the frame $\mathbf{I}$.
In the inference stage, $\mathbf{Q}$ from the decoder is passed directly through the \texttt{argmax} function to infer the final binary mask $\mathbf{Y}$. 
Prediction segmentation of video $\mathcal{I}$ without applying any post-processing techniques can be phrased as
\begin{equation}\label{Eq.12}
	\begin{split}
		\mathbf{Y}=\texttt{argmax}(\mathbf{Q})\in\{0,1\}^{{H}\times{W}}.
	\end{split}
\end{equation}

\section{Experiments}
\subsection{Experimental Setup}
\mt{Datasets.} 
We evaluate HFAN on three publicly available datasets with UVOS: DAVIS-16~\cite{DAVIS-16}, YouTube-Objects~\cite{YouTube-Objects} and Long-Videos~\cite{AFB-URR}. 
DAVIS-16~\cite{DAVIS-16} contains a total of 50 videos, including 30 videos for training and 20 videos for validation. YouTube-Objects~\cite{YouTube-Objects} includes 126 web videos divided into 10 categories with a total of more than 20,000 frames. Long-Videos~\cite{AFB-URR} consists of three videos, each of which contains about \textbf{2500} frames per video sequence.

\mt{Implementation Details.}
We utilize PyTorch~\cite{Pytorch} and MMSegmentation codebase~\cite{MMSegmentation} to implement our model and train on two NVIDIA V100 with a mini-batch size of 8 per GPU. 
To achieve a better trade-off between accuracy and speed, we choose lightweight MiT-b1 and middleweight MiT-b2 as the backbones rather than the better but larger MiT-b3 to MiT-b5~\cite{SegFormer}. 
Following~\cite{MATNet,RVOS,3DCSeg}, we pre-train our network on YouTube-VOS~\cite{YouTube-VOS} and fine-tune on the training set of DAVIS-16~\cite{DAVIS-16}. During training, we augment data online by random horizontal flipping, random resize with ratio $0.5\sim2.0$, and random cropping to $512\times{512}$. 
We use AdamW optimizer to pre-train for 160K iterations and fine-tune for 4K iterations. The learning rates of pre-training and fine-tuning are set to $6e-5$ with a \textit{poly} schedule and $1e-5$ with a \textit{fixed} schedule, respectively. 
For obtaining an elegant end-to-end model, we do not employ training tricks like auxiliary head loss and online hard example mining~\cite{OHEM}.
Moreover, no post-processing techniques (\eg, the widely used CRF~\cite{CRF}) are used in the inference phase.
All inference processes for the experiments are performed using a single V100. 
We report UVOS performance using two standard evaluation metrics recommended by~\cite{DAVIS-16}, \ie, region similarity $\mathcal{J}$ and boundary accuracy $\mathcal{F}$.

\subsection{Ablation Studies}\label{subsec:Ablation}
To quantify the effect of each fundamental component in HFAN, we perform an exhaustive ablation study on the DAVIS-16 val-set~\cite{DAVIS-16}. For the fairness of the ablation results, we do not perform any post-processing techniques.

\begin{table}[b]
\begin{minipage}{0.43\linewidth}
	\centering
	\caption{Ablation study for data input. All ablated versions utilize hierarchical architecture MiT-b1 as the backbone.}\label{tab:1}
	\resizebox{5.2cm}{!}{
		\begin{tabular}{c|cc|cc}
			\hline
			\textbf{Input} &  \textbf{$\mathcal{J}$ Mean $\uparrow$} & \textbf{$\Delta \mathcal{J}$} & \textbf{$\mathcal{F}$ Mean $\uparrow$} & \textbf{$\Delta \mathcal{F}$} \\
			\hline
			Image frame only & 79.1 &-3.9 & 79.8 &-3.5 \\
			Optical flow only & 77.9 &-5.1 & 76.5 &-6.8 \\
			\hline
			Baseline & 83.0 & - & 83.3 & - \\
			\hline
	\end{tabular}}
\end{minipage}
\hfill
\begin{minipage}{0.55\linewidth}
	\centering
	\caption{Ablation study for module. HFAN indicates a full model with integrated FAM and FAT modules.}\label{tab:2}
	\resizebox{6.7cm}{!}{
		\begin{tabular}{c|cc|cc|c}
			\hline
			\textbf{Variant} &  \textbf{$\mathcal{J}$ Mean $\uparrow$} & \textbf{$\Delta \mathcal{J}$} & \textbf{$\mathcal{F}$ Mean $\uparrow$} & \textbf{$\Delta \mathcal{F}$} &\textbf{FPS $\uparrow$} \\
			\hline
			Baseline & 83.0 & - & 83.3 & - &22.0\\
			\hline
			Baseline + FAM & 85.2 & +2.2 & 85.6 & +2.3 &21.0\\
			Baseline + FAT & 85.0 & +2.0 & 86.1 & +2.8 &21.4\\
			Baseline + HFAN & 86.2 & +3.2 & 87.1 & +3.8 &20.8\\
			\hline
	\end{tabular}}
\end{minipage}
\end{table}

\mt{Impact of Data Input.} 
To analyze the effect of appearance and motion features on performance, we first conduct an ablation study on the data input in Table~\ref{tab:1}. We adopt the video frame and corresponding optical flow as data inputs. A simple additive feature fusion approach is employed as the baseline. Compared to using a single input type, the baseline improves performance by providing richer appearance and motion cues. The ablation results illustrate that optical flow, which is deemed as the temporal consistency between video frames, requires the coaction of appearance features to achieve the desired effect.

\begin{table}[t]
\begin{minipage}{0.49\linewidth}
\centering
\caption{Ablation study on different backbones. Transformer-like and CNN-like versions are considered in the experimental ablation. For the test setup, SS / MS denotes single / multi-scale test.}\label{tab:3}
\resizebox{5.4cm}{!}{
\begin{tabular}{c|c|c|c|c}
			\hline
			\textbf{Backbone} & \textbf{Test Setup} & \textbf{$\mathcal{J}$ Mean $\uparrow$} & \textbf{$\mathcal{F}$ Mean $\uparrow$} & \textbf{FPS $\uparrow$} \\
			\hline
			\multirow{2}*{MiT-b0} & SS & 81.5 & 80.8 & 24.0 \\
			& MS & 83.4 & 82.3 & 3.4 \\
			\multirow{2}*{MiT-b1} & SS  & 86.2 & 87.1 & 20.8 \\
			~ & MS &87.1  &87.7 &2.5 \\
			\multirow{2}*{MiT-b2} & SS  & 86.8 & 88.2 & 14.4 \\
			~ & MS &88.0  &89.3 &1.4 \\
			\multirow{2}*{MiT-b3} & SS & 86.8 & 88.8 & 10.6 \\
			& MS & 88.2 & 90.0 & 1.0 \\
			\hline
			\multirow{2}*{Swin-Tiny} & SS & 86.0 & 87.3 & 12.8 \\
			~ & MS & 87.2 & 87.9 &1.1 \\
			\multirow{2}*{ResNet-101} & SS &86.6  &87.3  &12.4 \\
			~ & MS &87.3 &87.9 &1.3 \\
			\hline
\end{tabular}}
\end{minipage}
\hfill
\begin{minipage}{0.49\linewidth}
\centering
\caption{Ablation study on different input sizes and optical flow with MS test.}\label{tab:4}
\resizebox{5.4cm}{!}{
\begin{tabular}{|c|cc|cc|}
			\hline
			\multirow{2}*{\diagbox[dir=SE]{Size}{Method}} &\multicolumn{2}{c|}{RAFT} &\multicolumn{2}{c|}{PWCNet} \\
			& $\mathcal{J}$ Mean $\uparrow$ & $\mathcal{F}$ Mean $\uparrow$ &$\mathcal{J}$ Mean $\uparrow$ & $\mathcal{F}$ Mean $\uparrow$ \\
			\hline
			$384\times384$ &86.2 &86.6 &84.5 &84.7 \\
			$448\times448$ &86.9 &87.5 &85.3 &85.7 \\
			$480\times480$ &86.9 &87.6 &85.5 &85.9 \\
			$512\times512$ &87.1 &87.7 &85.7 &86.0 \\
			\hline
\end{tabular}}
\centering
\caption{Ablation study on Transformer-like and CNN-like network architectures. Ablated results are obtained in same setups (RAFT, $512\times512$, and MS test).}\label{tab:5}
\resizebox{5.4cm}{!}{
\begin{tabular}{|c|cc|cc|}
			\hline
			\multirow{2}*{\diagbox[dir=SE]{Backbone}{Method}} &\multicolumn{2}{c|}{MATNet + CRF} &\multicolumn{2}{c|}{Ours} \\
			& $\mathcal{J}$ Mean $\uparrow$ & $\mathcal{F}$ Mean $\uparrow$ &$\mathcal{J}$ Mean $\uparrow$ & $\mathcal{F}$ Mean $\uparrow$ \\
			\hline
			MiT-b1 &83.8 &82.6 &87.1 &87.7 \\
			MiT-b2 &84.7 &83.8 &88.0 &89.3 \\
			ResNet-101 &84.0 &82.9 &87.3 &87.9 \\
			\hline
\end{tabular}}
\end{minipage}
\end{table}

\begin{figure}[b]
	\centering
	\subfloat[Baseline]{\includegraphics[width=0.24\linewidth]{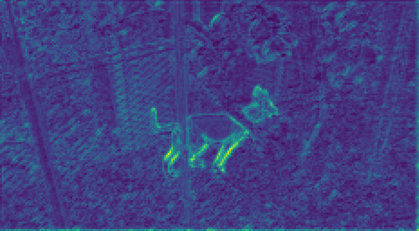}}
	\hfill
	\subfloat[+ FAM]{\includegraphics[width=0.24\linewidth]{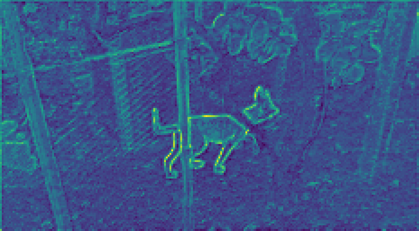}}
	\hfill
	\subfloat[+ FAT]{\includegraphics[width=0.24\linewidth]{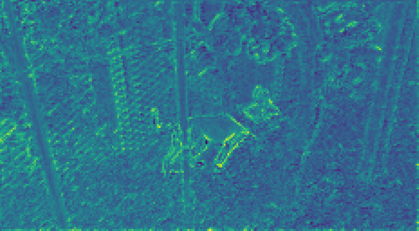}}
	\hfill
	\subfloat[+ HFAN]{\includegraphics[width=0.24\linewidth]{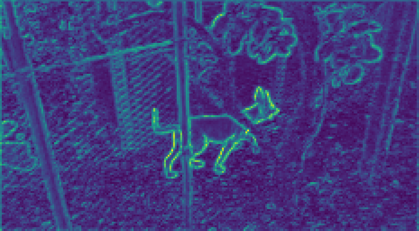}}
	\caption{Illustration of the first stage feature maps $\mathbf{U}_1$ from four ablated models.}\label{fig:ablation}
\end{figure}

\mt{Efficacy of Crucial Modules.}
When comparing our baseline with FAM, FAT, and HFAN, the results in Table~\ref{tab:2} reveal that HFAN is a superior aggregate of FAM and FAT. Specifically, FAM improves 2.2\% and 2.3\% in terms of $\mathcal{J}$ Mean and $\mathcal{F}$ Mean, respectively. FAT increases by 2.0\% on $\mathcal{J}$ Mean and 2.8\% $\mathcal{F}$ Mean. 
The best performance gains achieved by HFAN, which is implemented by combining FAM and FAT modules, further demonstrates the effectiveness of proposed approach. 
For aligning features of co-foreground objects in different modal images, HFAN achieves a simple way to correct shift differences between video frames and their corresponding optical flow features.
In addition, HFAN achieves adaptive selection in the feature fusion phase by learning feature adaptation weights. 
Fig.~\ref{fig:ablation} visualizes the ablated versions for Table~\ref{tab:2}. 
It can be found that FAM aligns the image and optical flow features to yield smoother and more refined object boundaries. 
Meanwhile, FAT enhances the image and optical flow features by adaptive transformation. 
Our HFAN inherits advantages of FAM and FAT, obtaining more finesse in the target region and removing a larger amount of noise in the non-target region.

\mt{Efficacy of Backbone.}
We investigate the effect of different backbone networks on accuracy and speed. The results of MiT-b0 to MiT-b3~\cite{SegFormer} are shown in Table~\ref{tab:3} (Note that we do not run experiments using MiT-b4 and MiT-b5 due to GPU memory limitations.). 
We find that the performance increases when enlarging the size of backbone networks. 
However, a larger network leads to a lower model efficiency and real-time speed. 
In addition, other types of backbone networks (\eg, Swin Transformer~\cite{Swin} and ResNet~\cite{ResNet}) also achieve competitive results. This adequately demonstrates the generality of the proposed method. 
Given the trade-off between the model size and performance, we choose MiT-b1 and MiT-b2 as the small and medium backbone networks for HFAN, respectively.

\mt{Effect of Image Size and Optical Flow.}
Low-resolution image inputs generally degrade the performance of the segmentation model, while the use of different optical flow estimation methods also affects the final segmentation results. To study the effects of image size and optical flow estimation methods on the proposed method, we explore four different image size inputs and two well-known optical flow estimation methods. The ablation results are shown in Table~\ref{tab:4}, and we can find that \textbf{1)} the proposed method still has good performance under the low-resolution condition; \textbf{2)} RAFT~\cite{RAFT} has better results than PWCNet~\cite{PWCNet} at the same resolution. The comprehensive analysis suggests that our method is not sensitive to the image resolution, while the optical flow estimation of different quality has a more obvious impact on the segmentation results.

\mt{Impact of Network Architecture.}
We further explore the impact of different network architectures on video segmentation methods. Table~\ref{tab:5} shows the ablation results of Transformer-like (MiT~\cite{SegFormer}) and CNN-like (ResNet~\cite{ResNet}) networks, and the analysis reveals that \textbf{1)} the performance ranking order of both methods (MATNet~\cite{MATNet} and ours) is MiT-b2 $>$ ResNet-101 $>$ MiT-b1, and \textbf{2)} the proposed method outperforms MATNet (Note that results of MATNet are obtained by the CRF post-processing technique, while our results are not.)  above 4.1\% in terms of $\mathcal{J}\&\mathcal{F}$ Mean for the same network architecture. The above ablation results show that the large Transformer-like MiT-b2~\cite{SegFormer} benefits from better visual perception and achieves better segmentation performance compared to CNN-like ResNet-101~\cite{ResNet}.

\subsection{Quantitative Results for UVOS}

\begin{table}[t]
	\centering
	\caption{Evaluation on DAVIS-16~\cite{DAVIS-16}. `small' and `medium' indicate that the backbone networks of HFAN are MiT-b1 and MiT-b2, respectively. `$\dag$' means that the optical flow is used. `PP' denotes post-processing. The three best scores are marked in \mkr{red}, \mkb{blue} and \mkg{green} for each metric, respectively. The inference speed (FPS) of each model contains all the necessary aspects for its generation of final results.}\label{tab:6}
	\resizebox{11.9cm}{!}{
		\begin{tabular}{r|c|c|ccc|ccc|c|c}
			\hline
			\multirow{2}*{\textbf{Method}~~~~~~~~~~~~~} & \multirow{2}*{\textbf{Publication}} & \multirow{2}*{\textbf{PP}} &\multicolumn{3}{c|}{\textbf{$\mathcal{J}$}} &\multicolumn{3}{c|}{\textbf{$\mathcal{F}$}} & \textbf{$\mathcal{J}$\&$\mathcal{F}$} & \multirow{2}*{\textbf{FPS $\uparrow$}} \\
			~ &~ &~ &\textbf{Mean $\uparrow$} &\textbf{Recall $\uparrow$} &\textbf{Decay $\downarrow$} &\textbf{Mean $\uparrow$} &\textbf{Recall $\uparrow$} &\textbf{Decay $\downarrow$} &\textbf{Mean $\uparrow$}  &~ \\
			\hline
			PDB~\cite{PDB} &ECCV 2018 &\checkmark &77.2 &90.1 &\mkg{0.9} &74.5 &84.4 &\mkr{-0.2} &75.9 &\mkg{20.0} \\
			UOVOS$^\dag$~\cite{UOVOS} &TIP 2019 &\checkmark &73.9 &88.5 &\mkb{0.6} &68.0 &80.6 &0.7 &71.0 &- \\
			LSMO$^\dag$~\cite{LSMO} &IJCV 2019 &\checkmark &78.2 &89.1 &4.1 &75.9 &84.7 &3.5 &77.1 &- \\
			MotAdapt$^\dag$~\cite{MotAdapt} &ICRA 2019 &\checkmark &77.2 &87.8 &5.0 &77.4 &84.4 &3.3 &77.3 &- \\
			AGS~\cite{AGS}  &CVPR 2019 &\checkmark &79.7 &91.1 &1.9 &77.4 &85.8 &1.6 &78.6 &1.7 \\
			AGNN~\cite{AGNN}  &ICCV 2019 &\checkmark&80.7 &94.0 &\mkr{0.0} &79.1 &90.5 &\mkb{0.0} &79.9 &1.9 \\
			COSNet~\cite{COSNet}  &CVPR 2019 &\checkmark &80.5 &93.1 &4.4 &79.5 &89.5 &5.0 &80.0 &2.2 \\
			AnDiff~\cite{AnDiff}  &ICCV 2019 & &81.7 &90.9 &2.2 &80.5 &85.1 &\mkg{0.6} &81.1 &2.8 \\
			PCSA~\cite{PCSA}  &AAAI 2020 &&78.1 &90.0 &4.4 &78.5 &88.1 &4.1 &78.3 &\mkr{110}\\
			EPO+$^\dag$~\cite{EPO+} &WACV 2020 &\checkmark &80.6 &95.2 &2.2 &75.5 &87.9 &2.4 &78.1 &- \\
			MATNet$^\dag$~\cite{MATNet}  &AAAI 2020 &\checkmark &82.4 &94.5 &3.8 &80.7 &90.2 &4.5 &81.5 &1.3 \\
			GraphMem~\cite{GraphMem} &ECCV 2020 &\checkmark &82.5 &94.3 &4.2 &81.2 &90.3 &5.6 &81.9 & 5.0\\
			DFNet~\cite{DFNet}  &ECCV 2020 &\checkmark &83.4 &94.4 &4.2 &81.8 &89.0 &3.7 &82.6 &3.6 \\
			3DCSeg~\cite{3DCSeg}  &BMVC 2020 &&84.2 &95.8 &7.4 &84.3 &92.4 &5.5 &84.2  &4.5 \\
			F2Net~\cite{F2Net}  &AAAI 2021 &&83.1 &95.7 &\mkr{0.0} &84.4 &92.3 &0.8 &83.7 &10.0 \\
			FSNet$^\dag$~\cite{FSNet}  &ICCV 2021 &\checkmark &83.4 &94.5 &3.2 &83.1 &90.2 &2.6 &83.3 &12.5 \\
			AMC-Net$^\dag$~\cite{AMC-Net} &ICCV 2021 &\checkmark &84.5 &\mkg{96.4} &2.8 &84.6 &{93.8} &2.5 &84.6 &- \\
			TransportNet$^\dag$~\cite{TransportNet} &ICCV 2021 &&84.5 &- &- &{85.0} &- &- &84.8 &3.6 \\
			RTNet$^\dag$~\cite{RTNet}  &CVPR 2021 &\checkmark &{85.6} &{96.1} &- &84.7 &{93.8} &- &{85.2} &- \\
			\hline
			\textbf{Ours-small$^\dag$(SS/MS)}&\multirow{3}*{-} &&86.2/\mkb{87.1} &\mkb{96.7}/\mkr{96.8} &4.6/4.8 &87.1/\mkg{87.7} &\mkr{95.5}/\mkg{95.3} &2.3/2.5 &86.7/\mkg{87.4}  &\mkb{20.8}/2.5 \\
			\textbf{Ours-medium$^\dag$(SS/MS)}&~ &&\mkg{86.8}/\mkr{88.0} &{96.1}/{96.2} &4.3/4.5 &\textcolor{blue}{88.2}/\textcolor{red}{89.3} &\mkg{95.3}/\mkb{95.4} &1.1/2.0 &\mkb{87.5}/\mkr{88.7}  &14.4/1.4 \\
			\hline
	\end{tabular}}
\end{table}

\begin{table}[t]
	\centering
	\caption{Evaluation on YouTube-Objects~\cite{YouTube-Objects}. The three best scores are marked in \mkr{red}, \mkb{blue}, and \mkg{green} for each object category over $\mathcal{J}$ Mean $\uparrow$.}\label{tab:7}
	\resizebox{11.9cm}{!}{
		\begin{tabular}{r|cccccccccccccc}
			\hline
			\multirow{2}*{\textbf{Method}~}  &MOTAdapt &LSMO &LVO &FSEG &PDB &SFL &AGS &COSNet &AGNN &MATNet &AMCNet &GraphMem &RTNet &\multirow{2}*{\textbf{Ours-small}}\\
			~ &\cite{MotAdapt} &\cite{LSMO} &\cite{LVO} &\cite{Fusionseg} &\cite{PDB} &\cite{Segflow} &\cite{AGS} &\cite{COSNet} &\cite{AGNN} &\cite{MATNet} &\cite{AMC-Net} &\cite{RTNet} &\cite{GraphMem} & \\
			\hline
			\textbf{Airplane} &77.2 &60.5 &\mkb{86.2} &81.7 &78.0 &65.6 &\mkr{87.7} &81.1 &81.1 &72.9 &78.9 &\mkg{86.1} &84.1 &84.7 \\
			\textbf{Bird} &42.2 &59.3 &\mkr{81.0} &63.8 &80.0 &65.4 &76.7 &75.7 &75.9 &77.5 &\mkb{80.9} &75.7 &\mkg{80.2} &80.0 \\
			\textbf{Boat} &49.3 &62.1 &68.5 &\mkr{72.3} &58.9 &59.9 &\mkb{72.2} &71.3 &70.7 &66.9 &67.4 &68.6 &70.1 &\mkg{72.0} \\
			\textbf{Car} &68.6 &72.3 &69.3 &74.9 &76.5 &64.0 &78.6 &77.6 &78.1 &79.0 &\mkb{82.0} &\mkr{82.4} &\mkg{79.5} &76.1 \\
			\textbf{Cat} &46.3 &66.3 &58.8 &68.4 &63.0 &58.9 &69.2 &66.5 &67.9 &\mkb{73.7} &69.0 &65.9 &\mkg{71.8} &\mkr{76.0} \\
			\textbf{Cow} &64.2 &67.9 &68.5 &68.0 &64.1 &51.2 &64.6 &69.8 &69.7 &67.4 &69.6 &\mkb{70.5} &\mkg{70.1} &\mkr{71.2} \\
			\textbf{Dog} &66.1 &70.0 &61.7 &69.4 &70.1 &54.1 &73.3 &76.8 &\mkr{77.4} &75.9 &75.8 &\mkb{77.1} &71.3 &\mkg{76.9} \\
			\textbf{Horse} &64.8 &65.4 &53.9 &60.4 &\mkg{67.6} &64.8 &64.4 &67.4 &67.3 &63.2 &63.0 &\mkr{72.2} &65.1 &\mkb{71.0} \\
			\textbf{Motorbike} &44.6 &55.5 &60.8 &62.7 &58.4 &52.6 &62.1 &\mkb{67.7} &\mkr{68.3} &62.6 &63.4 &63.8 &\mkg{64.6} &64.3 \\
			\textbf{Train} &42.3 &38.0 &\mkr{66.3} &\mkb{62.2} &35.3 &34.0 &48.2 &46.8 &47.8 &51.0 &57.8 &47.8 &53.3 &\mkg{61.4} \\
			\hline
			\textbf{\textit{Average}} &58.1 &64.3 &67.5 &68.4 &65.5 &57.1 &69.7 &70.5 &70.8 &69.0 &\mkg{71.1} &\mkb{71.4} &71.0 &\mkr{73.4} \\
			\hline
		\end{tabular}}
\end{table}

\begin{table}[t]
	\centering
	\caption{Evaluation on Long-Videos~\cite{AFB-URR}. The best results of SVOS and UVOS methods are marked in \underline{underline} and \textbf{bold}, respectively.}\label{tab:8}
	\resizebox{12cm}{!}{
		\begin{tabular}{r|c|ccc|ccc|c}
			\hline
			\multirow{2}*{\textbf{Method}~~~~~} &\multirow{2}*{\textbf{Supervision}} &\multicolumn{3}{c|}{\textbf{$\mathcal{J}$}} &\multicolumn{3}{c|}{\textbf{$\mathcal{F}$}} &\multirow{2}*{\textbf{$\mathcal{J}$\&$\mathcal{F}$ Mean $\uparrow$}} \\
			~ &~ &\textbf{Mean $\uparrow$} &\textbf{Recall $\uparrow$} &\textbf{Decay $\downarrow$} &\textbf{Mean $\uparrow$} &\textbf{Recall $\uparrow$} &\textbf{Decay $\downarrow$} &~ \\
			\hline
			RVOS~\cite{RVOS} &\multirow{4}*{SVOS} &10.2 &6.7 &13.0 &14.3 &11.7 &\underline{10.1} &12.2 \\
			A-GAME~\cite{A-GAME} &~  &50.0 &58.3 &39.6 &50.7 &58.3 &45.2 &50.3 \\
			STM~\cite{STM}  & &79.1 &88.3 &11.6 &79.5 &90.0 &15.4 &79.3 \\
			AFB-URR~\cite{AFB-URR} &~  &\underline{82.7} &\underline{91.7} &\underline{11.5} &\underline{83.8} &\underline{91.7} &13.9 &\underline{83.3} \\
			\hline
			3DCSeg~\cite{3DCSeg} &\multirow{5}*{UVOS} &34.2 &38.6 &11.6 &33.1 &28.1 &15.6 &33.7 \\
			MATNet~\cite{MATNet}  &~ &66.4 &73.7 &{10.9} &69.3 &77.2 &{10.6} &67.9 \\
			AGNN~\cite{AGNN} &~  &68.3 &77.2 &13.0 &68.6 &77.2 &16.6 &68.5 \\
			\textbf{Ours-small} &~  &{74.9} &{82.5} &14.8 &{76.1} &{86.0} &16.0 &{75.5} \\
			\textbf{Ours-medium} &~  &\textbf{80.2} &\textbf{91.2} &\textbf{9.4} &\textbf{83.2} &\textbf{96.5} &\textbf{7.1} &\textbf{81.7} \\
			\hline
	\end{tabular}}
\end{table}

\mt{DAVIS-16.}
We compare the proposed model HFAN with SOTA methods on the public benchmark DAVIS-16~\cite{DAVIS-16}. Table~\ref{tab:6} shows the quantitative results. 
Our method outperforms all existing SOTA models by a significant margin on DAVIS-16. 
Specifically, our HFAN-small scores 86.7\% $\mathcal{J}\&\mathcal{F}$ Mean and reaches 20.8 FPS in real-time speed. 
In contrast to RTNet~\cite{RTNet}, which employs both forward and backward optical flow and uses post-processing, HFAN-medium achieves 88.7\% $\mathcal{J}\&\mathcal{F}$ Mean using only forward optical flow without any post-processing techniques. 
Compared with previous methods~\cite{MATNet,FSNet,AMC-Net,TransportNet,RTNet} using optical flow, our method exhibits significant superiority in terms of inference speed and segmentation accuracy. The main reason is that the FAM and FAT modules in HFAN perform feature alignment and adaptation for unaligned cross-modal features, allowing the decoder to utilize a more accurate feature representation. 
Quantitative results of different metrics demonstrate that our method achieves a nice trade-off between accuracy and speed in the UVOS task.

\mt{YouTube-Objects.} To explore the universality of our proposed method to other video datasets, we perform validation experiments on the YouTube-Objects~\cite{YouTube-Objects} test set without further fine-tuning its training set. The quantitative results of 10 categories in this dataset are shown in Table~\ref{tab:7}. Our method HFAN-small dose not reach SOTA across all categories but has better stability than other comparative methods. The proposed method is 2.0\% higher than the second-best GraphMem~\cite{GraphMem} in terms of \textit{average} $\mathcal{J}$ Mean. For 10 different object categories, the proposed method achieves its balanced performance over various challenging (\eg, motion blur, occlusion, scale variation) video sequences. This is made possible by the sensible interaction of the proposed modules (FAM and FAT) for appearance and motion information.

\mt{Long-Videos.}
DAVIS~\cite{DAVIS-16} (\textbf{60+} frames per video sequence in average) only contains short-term video clips, while real-world videos tend to have more frames. To verify the performance of our HFAN in long-term video object segmentation, we evaluate it on the Long-Videos~\cite{AFB-URR} val-set (approximate \textbf{2500} frames per video sequence). Table~\ref{tab:8} shows the results under two types of supervision, SVOS and UVOS. By further observation, we can find that the proposed HFAN-medium has obtained the best result, achieving 81.7$\%$ over $\mathcal{J}$\&$\mathcal{F}$ Mean under the UVOS setting. Compared with the second-best method AGNN~\cite{AGNN}, our small model obtains an improvement of 7.0\% on $\mathcal{J}$\&$\mathcal{F}$ Mean. 
Meanwhile, HFAN-medium achieves appealing results compared to SVOS methods. 
The results show that the temporal consistency provided by optical flow is also effective for long-term video object segmentation.

\subsection{Quantitative Results for VSOD}
The additional task VSOD, like UVOS, does not require first-frame annotation. To verify the performance of the proposed method on the VSOD setting, we perform a quantitative comparison with eight SOTA models on DAVIS~\cite{DAVIS-16}.

\begin{table}[t]
	\centering
	\caption{Evaluation on DAVIS~\cite{DAVIS-16} for VSOD. The best scores are marked in \textbf{bold}.}\label{tab:9}
		\resizebox{12cm}{!}{
			\begin{tabular}{r|cccccccccc}
				\hline
				\multirow{2}*{\textbf{Method}} &FGRN &LTSI &RCR &MBN &SSAV &PCSA &DCFNet &FSNet &\multirow{2}*{\textbf{Ours-small}} &\multirow{2}*{\textbf{Ours-medium}}  \\
				&\cite{FGRN} &\cite{LTSI} &\cite{RCR} &~\cite{MBN} &~\cite{SSAV} &~\cite{PCSA} &~\cite{DCFNet} &~\cite{FSNet} & & \\
				\hline
				\textbf{$S_\alpha \uparrow$} &0.838 &0.876 &0.886 &0.887 &0.893 &0.902 &0.914 &0.920 &0.934 &\textbf{0.938} \\
				\textbf{$E_{\xi}^{max} \uparrow$} &0.917 &0.957 &0.947 &0.966 &0.948 &0.961 &- &0.970 &\textbf{0.983} &\textbf{0.983} \\
				\textbf{$F_{\beta}^{max} \uparrow$} &0.783 &0.850 &0.848 &0.862 &0.861 &0.880 &0.900 &0.907 &0.929 &\textbf{0.935} \\
				\textbf{$MAE \downarrow$} &0.043 &0.034 &0.027 &0.031 &0.028 &0.022 &0.016 &0.020 &0.009 &\textbf{0.008} \\
				\hline
		\end{tabular}}
\end{table}

\mt{Metrics.}
We employ four widely-used evaluation metrics including structure-measure $S_\alpha$ ($\alpha=0.5$), max enhanced-alignment measure $E_{\xi}^{max}$, max F-measure $F_{\beta}^{max}$ ($\beta^2=0.3$), and mean absolute error ($MAE$).

\mt{Results.}
As shown in Table~\ref{tab:9}, our HFAN outperforms all SOTA models. 
In particular, compared with DCFNet~\cite{DCFNet}, $S_\alpha$ and $F_{\beta}^{max}$ are improved by $\sim$2\% and $\sim$3\%, respectively. 
Compared to FSNet~\cite{FSNet}, HFAN achieves $>$1.3\% performance gains on $S_\alpha$, $E_{\xi}^{max}$ and $F_{\beta}^{max}$, and reduces $MAE$ by a factor of two.
This significantly proves the adaptability of our method to similar tasks.

\subsection{Qualitative Results}

\begin{figure}[t]
	\centering
	\includegraphics[width=1.0\linewidth]{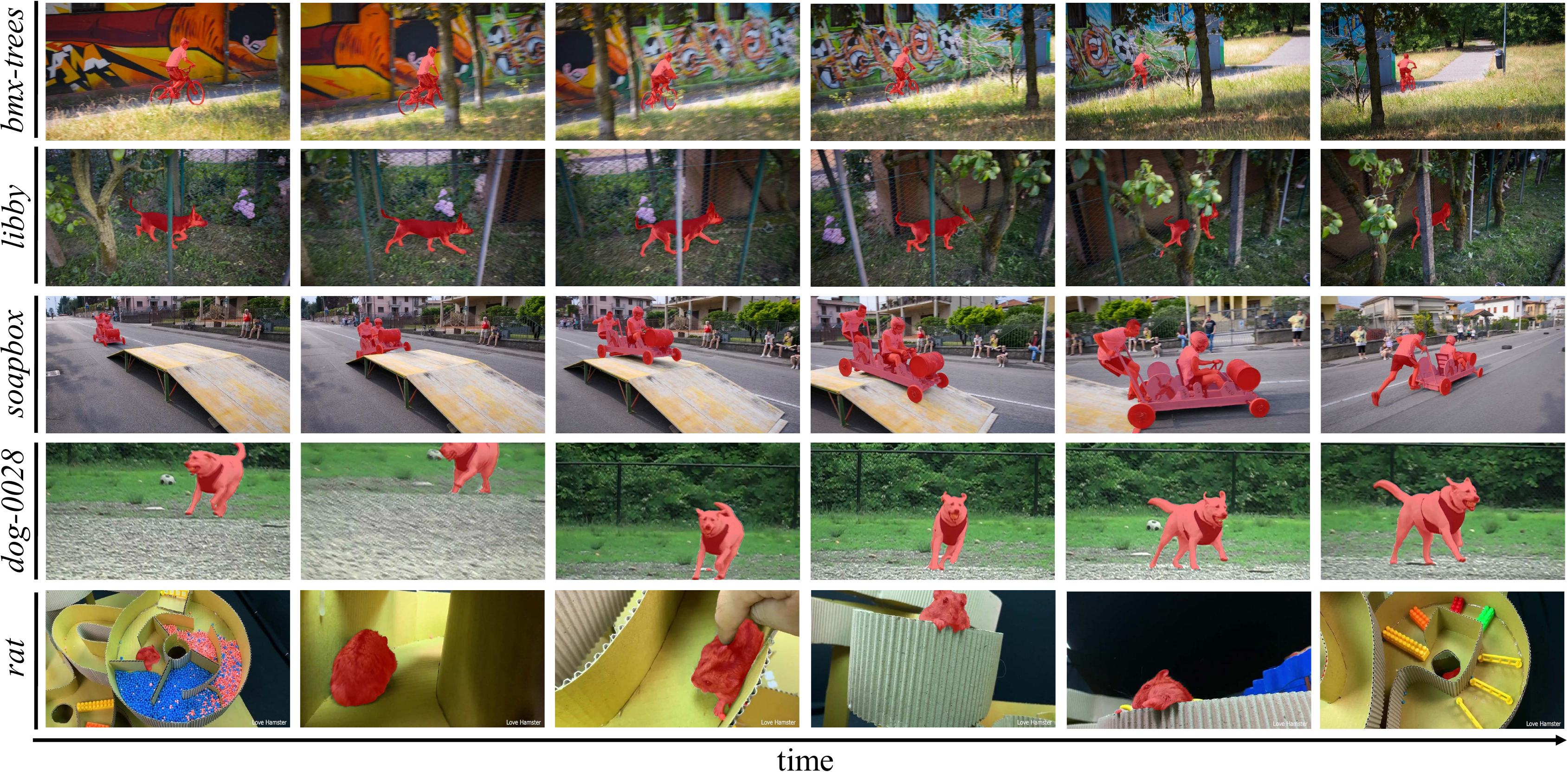}
	\caption{Qualitative results on three challenging video clips over time. From top to bottom: \textit{bmx-trees}, \textit{libby} and \textit{soapbox} from DAVIS-16~\cite{DAVIS-16}, \textit{dog-0028} from YouTube-Objects~\cite{YouTube-Objects}, and \textit{rat} from Long-Videos~\cite{AFB-URR}.}
	\label{fig:vis}
\end{figure}

Fig. \ref{fig:vis} shows qualitative results of our HFAN model. We select five videos from DAVIS-16~\cite{DAVIS-16}, YouTube-Objects~\cite{YouTube-Objects} and Long-Videos~\cite{AFB-URR} test sets. These videos consist of several challenging frame sequences (\eg, fast motion, scale variation, interacting objects and occlusion). As shown in the top two rows, our method yields desirable results for dynamic, similar, and complex backgrounds. Moreover, our proposed model has an accurate prediction for the occlusion boundary. In the third and fourth rows, satisfactory segmentation results are acquired in the presence of large-scale variation and object interaction cases.

\section{Conclusion}
We present a hierarchical feature alignment network, termed as HFAN, for addressing the contextual mismatch between appearance and motion features in the UVOS task. Firstly, to address the mismatch of primary object positions between video frames and their corresponding optical flows, our proposed FAM module relies on sharing primary objects in images across modalities to align appearance and motion features. Subsequently, for tackling the modal mismatch problem between aligned feature maps, the FAT module is designed to construct a feature adaptation weight to automatically enhance cross-modal features. With the alignment and adaptation of appearance and motion features achieved by FAM and FAT, HFAN could achieve a more accurate object segmentation. Experimental results show that the proposed method achieves SOTA performance in the unsupervised video object segmentation task.

\subsubsection{Acknowledgment.}
This work was supported by the National Natural Science Foundation of China (No. 62102182 and 61976116), Natural Science Foundation of Jiangsu Province (No. BK20210327), and Fundamental Research Funds for the Central Universities (No. 30920021135).

\bibliographystyle{splncs04}
\bibliography{egbib}

\begin{thebibliography}{10}
\providecommand{\url}[1]{\texttt{#1}}
\providecommand{\urlprefix}{URL }
\providecommand{\doi}[1]{https://doi.org/#1}

\bibitem{EPO+}
Akhter, I., Ali, M., Faisal, M., Hartley, R.: Epo-net: Exploiting geometric
  constraints on dense trajectories for motion saliency. In: WACV (2020)

\bibitem{LTSI}
Chen, C., Wang, G., Peng, C., Zhang, X., Qin, H.: Improved robust video
  saliency detection based on long-term spatial-temporal information. TIP
  (2019)

\bibitem{I2CRC}
Chen, T., Yao, Y., Zhang, L., Wang, Q., Xie, G., Shen, F.: Saliency guided
  inter-and intra-class relation constraints for weakly supervised semantic
  segmentation. TMM  (2022)

\bibitem{AlignDet}
Chen, Y., Han, C., Wang, N., Zhang, Z.: Revisiting feature alignment for
  one-stage object detection. arXiv preprint arXiv:1908.01570  (2019)

\bibitem{MiVOS}
Cheng, H.K., Tai, Y.W., Tang, C.K.: Modular interactive video object
  segmentation: Interaction-to-mask, propagation and difference-aware fusion.
  In: CVPR (2021)

\bibitem{Segflow}
Cheng, J., Tsai, Y.H., Wang, S., Yang, M.H.: Segflow: Joint learning for video
  object segmentation and optical flow. In: ICCV (2017)

\bibitem{MMSegmentation}
Contributors, M.: {MMSegmentation}: Openmmlab semantic segmentation toolbox and
  benchmark. \url{https://github.com/open-mmlab/mmsegmentation} (2020)

\bibitem{SSTVOS}
Duke, B., Ahmed, A., Wolf, C., Aarabi, P., Taylor, G.W.: Sstvos: Sparse
  spatiotemporal transformers for video object segmentation. In: CVPR (2021)

\bibitem{SSAV}
Fan, D.P., Wang, W., Cheng, M.M., Shen, J.: Shifting more attention to video
  salient object detection. In: CVPR (2019)

\bibitem{POLP}
Giordano, D., Murabito, F., Palazzo, S., Spampinato, C.: Superpixel-based video
  object segmentation using perceptual organization and location prior. In:
  CVPR (2015)

\bibitem{PCSA}
Gu, Y., Wang, L., Wang, Z., Liu, Y., Cheng, M.M., Lu, S.P.: Pyramid constrained
  self-attention network for fast video salient object detection. In: AAAI
  (2020)

\bibitem{DMFBS}
Han, B., Davis, L.S.: Density-based multifeature background subtraction with
  support vector machine. TPAMI  (2011)

\bibitem{S2ANet}
Han, J., Ding, J., Li, J., Xia, G.S.: Align deep features for oriented object
  detection. TGRS  (2021)

\bibitem{ResNet}
He, K., Zhang, X., Ren, S., Sun, J.: Deep residual learning for image
  recognition. In: CVPR (2016)

\bibitem{GIVOS}
Heo, Y., Koh, Y.J., Kim, C.S.: Guided interactive video object segmentation
  using reliability-based attention maps. In: CVPR (2021)

\bibitem{FaPN}
Huang, S., Lu, Z., Cheng, R., He, C.: Fapn: Feature-aligned pyramid network for
  dense image prediction. In: ICCV (2021)

\bibitem{AlignSeg}
Huang, Z., Wei, Y., Wang, X., Shi, H., Liu, W., Huang, T.S.: Alignseg:
  Feature-aligned segmentation networks. TPAMI  (2021)

\bibitem{CSTM}
Hui, T., Huang, S., Liu, S., Ding, Z., Li, G., Wang, W., Han, J., Wang, F.:
  Collaborative spatial-temporal modeling for language-queried video actor
  segmentation. In: CVPR (2021)

\bibitem{FlowNet-2.0}
Ilg, E., Mayer, N., Saikia, T., Keuper, M., Dosovitskiy, A., Brox, T.: Flownet
  2.0: Evolution of optical flow estimation with deep networks. In: CVPR (2017)

\bibitem{BatchNorm}
Ioffe, S., Szegedy, C.: Batch normalization: Accelerating deep network training
  by reducing internal covariate shift. In: ICML (2015)

\bibitem{Fusionseg}
Jain, S.D., Xiong, B., Grauman, K.: Fusionseg: Learning to combine motion and
  appearance for fully automatic segmentation of generic objects in videos. In:
  CVPR (2017)

\bibitem{FSNet}
Ji, G.P., Fu, K., Wu, Z., Fan, D.P., Shen, J., Shao, L.: Full-duplex strategy
  for video object segmentation. In: ICCV (2021)

\bibitem{A-GAME}
Johnander, J., Danelljan, M., Brissman, E., Khan, F.S., Felsberg, M.: A
  generative appearance model for end-to-end video object segmentation. In:
  CVPR (2019)

\bibitem{RefVOS}
Khoreva, A., Rohrbach, A., Schiele, B.: Video object segmentation with language
  referring expressions. In: ACCV (2018)

\bibitem{CRF}
Kr{\"a}henb{\"u}hl, P., Koltun, V.: Efficient inference in fully connected crfs
  with gaussian edge potentials. In: NeurIPS (2011)

\bibitem{FGVI}
Lao, D., Zhu, P., Wonka, P., Sundaramoorthi, G.: Flow-guided video inpainting
  with scene templates. In: ICCV (2021)

\bibitem{FGRN}
Li, G., Xie, Y., Wei, T., Wang, K., Lin, L.: Flow guided recurrent neural
  encoder for video salient object detection. In: CVPR (2018)

\bibitem{MGA}
Li, H., Chen, G., Li, G., Yu, Y.: Motion guided attention for video salient
  object detection. In: ICCV (2019)

\bibitem{MBN}
Li, S., Seybold, B., Vorobyov, A., Lei, X., Kuo, C.C.J.: Unsupervised video
  object segmentation with motion-based bilateral networks. In: ECCV (2018)

\bibitem{SKNet}
Li, X., Wang, W., Hu, X., Yang, J.: Selective kernel networks. In: CVPR (2019)

\bibitem{SFNet}
Li, X., You, A., Zhu, Z., Zhao, H., Yang, M., Yang, K., Tan, S., Tong, Y.:
  Semantic flow for fast and accurate scene parsing. In: ECCV (2020)

\bibitem{AFB-URR}
Liang, Y., Li, X., Jafari, N., Chen, Q.: Video object segmentation with
  adaptive feature bank and uncertain-region refinement. In: NeurIPS (2020)

\bibitem{FPN}
Lin, T.Y., Doll{\'a}r, P., Girshick, R., He, K., Hariharan, B., Belongie, S.:
  Feature pyramid networks for object detection. In: CVPR (2017)

\bibitem{F2Net}
Liu, D., Yu, D., Wang, C., Zhou, P.: F2net: Learning to focus on the foreground
  for unsupervised video object segmentation. In: AAAI (2021)

\bibitem{Swin}
Liu, Z., Lin, Y., Cao, Y., Hu, H., Wei, Y., Zhang, Z., Lin, S., Guo, B.: Swin
  transformer: Hierarchical vision transformer using shifted windows. In: ICCV
  (2021)

\bibitem{GraphMem}
Lu, X., Wang, W., Danelljan, M., Zhou, T., Shen, J., Van~Gool, L.: Video object
  segmentation with episodic graph memory networks. In: ECCV (2020)

\bibitem{COSNet}
Lu, X., Wang, W., Ma, C., Shen, J., Shao, L., Porikli, F.: See more, know more:
  Unsupervised video object segmentation with co-attention siamese networks.
  In: CVPR (2019)

\bibitem{3DCSeg}
Mahadevan, S., Athar, A., O{\v{s}}ep, A., Hennen, S., Leal-Taix{\'e}, L.,
  Leibe, B.: Making a case for 3d convolutions for object segmentation in
  videos. In: BMVC (2020)

\bibitem{JOINT}
Mao, Y., Wang, N., Zhou, W., Li, H.: Joint inductive and transductive learning
  for video object segmentation. In: ICCV (2021)

\bibitem{PGFA}
Miao, J., Wu, Y., Liu, P., Ding, Y., Yang, Y.: Pose-guided feature alignment
  for occluded person re-identification. In: ICCV (2019)

\bibitem{ReLU}
Nair, V., Hinton, G.E.: Rectified linear units improve restricted boltzmann
  machines. In: ICML (2010)

\bibitem{SMO}
Ochs, P., Malik, J., Brox, T.: Segmentation of moving objects by long term
  video analysis. TPAMI  (2013)

\bibitem{IVOS}
Oh, S.W., Lee, J.Y., Xu, N., Kim, S.J.: Fast user-guided video object
  segmentation by interaction-and-propagation networks. In: CVPR (2019)

\bibitem{STM}
Oh, S.W., Lee, J.Y., Xu, N., Kim, S.J.: Video object segmentation using
  space-time memory networks. In: ICCV (2019)

\bibitem{FOSUV}
Papazoglou, A., Ferrari, V.: Fast object segmentation in unconstrained video.
  In: ICCV (2013)

\bibitem{Pytorch}
Paszke, A., Gross, S., Massa, F., Lerer, A., Bradbury, J., Chanan, G., Killeen,
  T., Lin, Z., Gimelshein, N., Antiga, L., Desmaison, A., Kopf, A., Yang, E.,
  DeVito, Z., Raison, M., Tejani, A., Chilamkurthy, S., Steiner, B., Fang, L.,
  Bai, J., Chintala, S.: Pytorch: An imperative style, high-performance deep
  learning library. In: NeurIPS (2019)

\bibitem{DAVIS-16}
Perazzi, F., Pont-Tuset, J., McWilliams, B., Van~Gool, L., Gross, M.,
  Sorkine-Hornung, A.: A benchmark dataset and evaluation methodology for video
  object segmentation. In: CVPR (2016)

\bibitem{FCOP}
Perazzi, F., Wang, O., Gross, M., Sorkine-Hornung, A.: Fully connected object
  proposals for video segmentation. In: ICCV (2015)

\bibitem{YouTube-Objects}
Prest, A., Leistner, C., Civera, J., Schmid, C., Ferrari, V.: Learning object
  class detectors from weakly annotated video. In: CVPR (2012)

\bibitem{RTNet}
Ren, S., Liu, W., Liu, Y., Chen, H., Han, G., He, S.: Reciprocal
  transformations for unsupervised video object segmentation. In: CVPR (2021)

\bibitem{URVOS}
Seo, S., Lee, J.Y., Han, B.: Urvos: Unified referring video object segmentation
  network with a large-scale benchmark. In: ECCV (2020)

\bibitem{HMMN}
Seong, H., Oh, S.W., Lee, J.Y., Lee, S., Lee, S., Kim, E.: Hierarchical memory
  matching network for video object segmentation. In: ICCV (2021)

\bibitem{OHEM}
Shrivastava, A., Gupta, A., Girshick, R.: Training region-based object
  detectors with online hard example mining. In: CVPR (2016)

\bibitem{MotAdapt}
Siam, M., Jiang, C., Lu, S., Petrich, L., Gamal, M., Elhoseiny, M., Jagersand,
  M.: Video object segmentation using teacher-student adaptation in a human
  robot interaction (hri) setting. In: ICRA (2019)

\bibitem{PDB}
Song, H., Wang, W., Zhao, S., Shen, J., Lam, K.M.: Pyramid dilated deeper
  convlstm for video salient object detection. In: ECCV (2018)

\bibitem{PWCNet}
Sun, D., Yang, X., Liu, M.Y., Kautz, J.: Pwc-net: Cnns for optical flow using
  pyramid, warping, and cost volume. In: CVPR (2018)

\bibitem{RAFT}
Teed, Z., Deng, J.: Raft: Recurrent all-pairs field transforms for optical
  flow. In: ECCV (2020)

\bibitem{LVO}
Tokmakov, P., Alahari, K., Schmid, C.: Learning video object segmentation with
  visual memory. In: ICCV (2017)

\bibitem{LSMO}
Tokmakov, P., Schmid, C., Alahari, K.: Learning to segment moving objects. IJCV
   (2019)

\bibitem{Attention}
Vaswani, A., Shazeer, N., Parmar, N., Uszkoreit, J., Jones, L., Gomez, A.N.,
  Kaiser, {\L}., Polosukhin, I.: Attention is all you need. In: NeurIPS (2017)

\bibitem{RVOS}
Ventura, C., Bellver, M., Girbau, A., Salvador, A., Marques, F., Giro-i Nieto,
  X.: Rvos: End-to-end recurrent network for video object segmentation. In:
  CVPR (2019)

\bibitem{JPFA}
Wang, G., Zhang, T., Cheng, J., Liu, S., Yang, Y., Hou, Z.: Rgb-infrared
  cross-modality person re-identification via joint pixel and feature
  alignment. In: ICCV (2019)

\bibitem{AGNN}
Wang, W., Lu, X., Shen, J., Crandall, D.J., Shao, L.: Zero-shot video object
  segmentation via attentive graph neural networks. In: ICCV (2019)

\bibitem{SAG}
Wang, W., Shen, J., Porikli, F.: Saliency-aware geodesic video object
  segmentation. In: CVPR (2015)

\bibitem{AGS}
Wang, W., Song, H., Zhao, S., Shen, J., Zhao, S., Hoi, S.C., Ling, H.: Learning
  unsupervised video object segmentation through visual attention. In: CVPR
  (2019)

\bibitem{SegFormer}
Xie, E., Wang, W., Yu, Z., Anandkumar, A., Alvarez, J.M., Luo, P.: Segformer:
  Simple and efficient design for semantic segmentation with transformers. In:
  NeurIPS (2021)

\bibitem{YouTube-VOS}
Xu, N., Yang, L., Fan, Y., Yang, J., Yue, D., Liang, Y., Price, B., Cohen, S.,
  Huang, T.: Youtube-vos: Sequence-to-sequence video object segmentation. In:
  ECCV (2018)

\bibitem{RCR}
Yan, P., Li, G., Xie, Y., Li, Z., Wang, C., Chen, T., Lin, L.: Semi-supervised
  video salient object detection using pseudo-labels. In: ICCV (2019)

\bibitem{AMC-Net}
Yang, S., Zhang, L., Qi, J., Lu, H., Wang, S., Zhang, X.: Learning
  motion-appearance co-attention for zero-shot video object segmentation. In:
  ICCV (2021)

\bibitem{AnDiff}
Yang, Z., Wang, Q., Bertinetto, L., Hu, W., Bai, S., Torr, P.H.: Anchor
  diffusion for unsupervised video object segmentation. In: ICCV (2019)

\bibitem{NSROM}
Yao, Y., Chen, T., Xie, G.S., Zhang, C., Shen, F., Wu, Q., Tang, Z., Zhang, J.:
  Non-salient region object mining for weakly supervised semantic segmentation.
  In: CVPR (2021)

\bibitem{Jo-SRC}
Yao, Y., Sun, Z., Zhang, C., Shen, F., Wu, Q., Zhang, J., Tang, Z.: Jo-src: A
  contrastive approach for combating noisy labels. In: CVPR (2021)

\bibitem{EDI}
Yao, Y., Zhang, J., Shen, F., Hua, X., Xu, J., Tang, Z.: Exploiting web images
  for dataset construction: A domain robust approach. TMM  (2017)

\bibitem{OCR}
Yuan, Y., Chen, X., Wang, J.: Object-contextual representations for semantic
  segmentation. In: ECCV (2020)

\bibitem{TransportNet}
Zhang, K., Zhao, Z., Liu, D., Liu, Q., Liu, B.: Deep transport network for
  unsupervised video object segmentation. In: ICCV (2021)

\bibitem{DCFNet}
Zhang, M., Liu, J., Wang, Y., Piao, Y., Yao, S., Ji, W., Li, J., Lu, H., Luo,
  Z.: Dynamic context-sensitive filtering network for video salient object
  detection. In: ICCV (2021)

\bibitem{DFNet}
Zhen, M., Li, S., Zhou, L., Shang, J., Feng, H., Fang, T., Quan, L.: Learning
  discriminative feature with crf for unsupervised video object segmentation.
  In: ECCV (2020)

\bibitem{TAODA}
Zhou, T., Li, J., Li, X., Shao, L.: Target-aware object discovery and
  association for unsupervised video multi-object segmentation. In: CVPR (2021)

\bibitem{MATNet}
Zhou, T., Wang, S., Zhou, Y., Yao, Y., Li, J., Shao, L.: Motion-attentive
  transition for zero-shot video object segmentation. In: AAAI (2020)

\bibitem{UOVOS}
Zhuo, T., Cheng, Z., Zhang, P., Wong, Y., Kankanhalli, M.: Unsupervised online
  video object segmentation with motion property understanding. TIP  (2019)

\end{thebibliography}
\end{document}